\documentclass[preprints,article,accept,pdftex,moreauthors]{Definitions/mdpi} 
\usepackage[final]{changes}
\definechangesauthor[name={Chaofan}, color=orange]{CF.}
\usepackage{amsmath,amsfonts}
\usepackage{algorithmic}
\usepackage[ruled,linesnumbered,vlined]{algorithm2e}
\usepackage{array}
\usepackage[caption=false,font=normalsize,labelfont=sf,textfont=sf]{subfig}
\usepackage{textcomp}
\usepackage{stfloats}
\usepackage{url}
\usepackage{verbatim}
\usepackage{graphicx}
\usepackage[multiple]{footmisc}
\usepackage{multirow}
\usepackage{authblk}

\firstpage{1} 
\pubvolume{1}
\issuenum{1}
\articlenumber{0}
\pubyear{2022}
\copyrightyear{2022}
\datereceived{} 
\dateaccepted{} 
\datepublished{} 
\hreflink{https://doi.org/} 

\Title{Pyramidal Predictive Network: A Model for Visual-frame Prediction Based on Predictive Coding Theory}

\TitleCitation{Title}

\Author{Chaofan Ling $^{1}$, Junpei Zhong $^{2,}$* and Weihua Li $^{1,}$*}

\AuthorNames{Firstname Lastname, Firstname Lastname and Firstname Lastname}

\AuthorCitation{Ling, C.; Zhong, J.; Wei, H.}

\address{%
$^{1}$ \quad South China University of Technology, Guangzhou, China.; wichaofan@mail.scut.edu.cn (C.L.); whlee@scut.edu.cn (W.L.)\\
$^{2}$ \quad The Hong Kong Polytechnic University, KLN, Hong Kong.; joni.zhong@polyu.edu.hk}

\corres{Correspondence: joni.zhong@polyu.edu.hk (J.Z.) whlee@scut.edu.cn (W.L.) }

\abstract{\added{Visual-frame prediction is a pixel-dense prediction task that infers future frames from past frames. Lacking of appearance details, low prediction accuracy and high computational overhead are still major problems with current models or methods. In this paper, we propose a novel neural network model inspired by the well-known predictive coding theory to deal with the problems. Predictive coding provides an interesting and reliable computational framework, which will be combined with other theories such as the cerebral cortex at different level oscillates at different frequencies, to design an efficient and reliable predictive network model for visual-frame prediction. }  
\deleted{Inspired by the well-known predictive coding theory in cognitive science, we propose a novel neural network model for the task of visual-frame prediction.  In this paper, our main work is to combine the theoretical framework of predictive coding and deep learning architectures, to design an efficient and reliable predictive network model for visual-frame prediction.} 
Specifically, the model is composed of a series of recurrent and convolutional units forming the top-down and bottom-up streams, respectively. 
\deleted{It learns to predict future frames in a visual sequence, with ConvLSTMs on each layer in the network making local prediction from top to down. The main innovation of our model is that}
The update frequency of neural units on each of the layer decreases with the increasing of network levels, which results in \added{neurons of higher-level can capture information in longer time dimensions}.
\deleted{the model appears like a pyramid from the perspective of time dimension, so we call it the Pyramid Predictive Network (PPNet). Particularly, this pyramid-like design is consistent to the neuronal activities in the neuroscience findings involved in the predictive coding framework.}
According to the experimental results, this model shows better compactness and comparable predictive performance with existing works, implying lower computational cost and higher prediction accuracy. Code is available at \url{https://github.com/Ling-CF/PPNet}.}


\begin{document}


\section{Introduction}
The idea that brains are essentially prediction machines is one of the unified theories in cognitive science. It holds that brain functions, such as perception, motor control and memory, are all formed and modulated by prediction. Particularly, it also forms a sensorimotor framework (predictive coding) for understanding how human takes an action based on predictions. It proposes that the most functions in the brain follow a predictive manner, which is expressed by our brain's internal model. Therefore, the brain can continuously predict and form our perception, based on which we can also execute the motor actions. Such internal predictive model, shaped by the neurons' representation, is also always learning and updating itself in order to predict the changing environment better. This idea, if it is properly implemented by learning architectures, will be also useful in practical applications such as video-frame prediction.

\begin{figure}[]
	\centering{\includegraphics[width=4in]{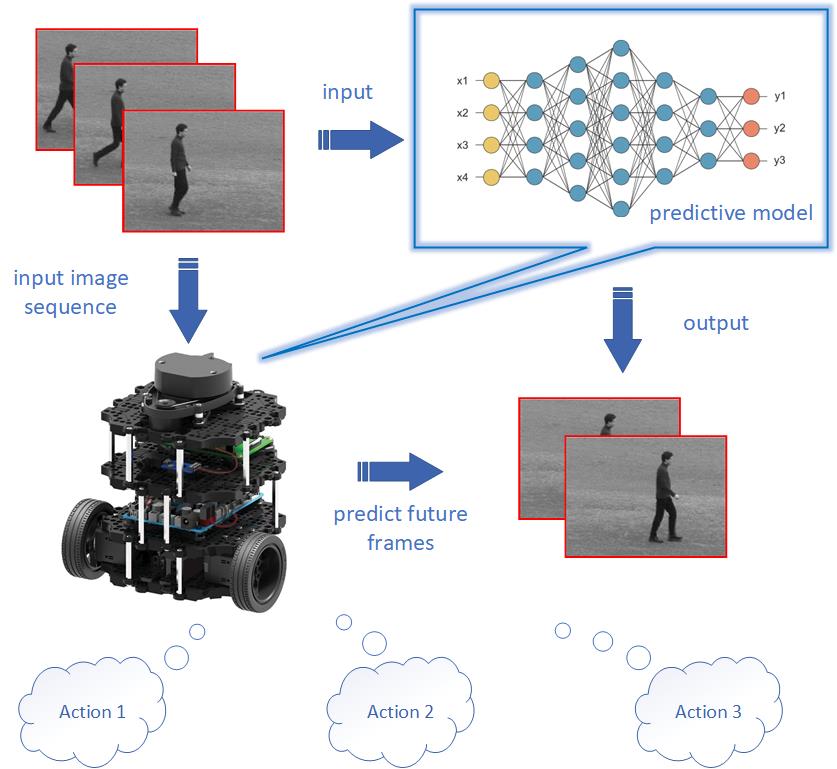}}
	\caption{A robot prediction system. By giving the context image sequences, the robot will predict future frames with a predictive model and make corresponding actions beforehand based on the predictions. }
	\label{fig:robot system}
\end{figure}

The so-called video-frame prediction is to predict the future of a visual frame based on the given context frames. From the perspective of applications, being able to predict the future  is of great significance. Adaptive systems that can predict how future scenes can be unfolded based on the internal model learned by the context will offer possibilities. For example, the predictive ability endows robots to foresee the future and even understand human's intention by analysing their movements, actions, etc., to make correct actions ahead of time (Figure \ref{fig:robot system}). \deleted{With the predictive ability,} Self-driving cars can anticipate the forthcoming situations and make judgments beforehand \cite{morris2008learning}. Moreover, there are a number of applications such as anticipating activities and events \cite{kitani2012activity}, long-term planning, prediction of pedestrian trajectories in traffic \cite{bhattacharyya2018long}, precipitation forecasting \cite{shi2017deep} and so on. \added{With the predictive ability, applications can become more efficient, they can foresee a changing future and react accordingly in advance, making their behavior smoother and more energy efficient. For different domains, the methods used may have some subtle differences (For instance, in the field of autonomous driving, the scene may be more complex, so we need a larger and deeper neural network, or other effective preprocessing or post-processing methods), but the overall framework of the model should be unchanged. }

Building on the success of deep learning, although a number of models or methods for visual-frame prediction have been proposed, the accuracy of predicted frames is still far from the requirements. This problem is more severe when performing long-term prediction or predicting visual sequences with large changes between frames. Besides, in view of the large computational overhead of existing models, making the model calculate in a more efficient way to promote the implementation of algorithm is another promising direction.

Therefore, in this work, we proposed to combine the theoretical framework of predictive coding and deep learning methods, to design a more efficient network model for the task of visual-frame prediction. This cognitive-inspired framework is a hierarchical processing model, which mimics the hierarchical processing structure of the cerebral cortex.  One of the main advantages of such a predictive coding model is that the internal model is updated by a combination of bottom-up and top-down information stream instead of just relying on outside information. This provides a possible framework for simulating and predicting the environment, which is also the essence that early works tried to implement as computational models \cite{softky1996unsupervised, deco2001predictive}.

The main contributions of this work are as follows: 1) We propose and construct a novel artificial neural network model, this model is a hierarchical network, which we call it the pyramidal predictive network (PPNet). It was modified on the basis of a generic framework proposed by “predictive coding''. As the name suggests, the update rating of neurons reduces with the increasing of the network level, which mimics the phenomenon of lower oscillations in the higher area of the visual cortex, and makes the model encodes information at various temporal and spatial scales as a result. 2) The loss function was improved to match the video prediction task. Inspired by the attention mechanism (for example, when the prediction differs greatly from the reality, the brain will react more strongly), we introduced the method of adaptive weight in the loss function, that is, the greater the prediction error, the greater the weight was given.  According to the results, the proposed methods do get a better prediction using less computational cost with a more compact and more time-dependent architecture. Later we will introduce our methods and the basis in detail.

The rest of this article is organized as follows: First, Section 2 reviews the related work about “Predictive Brains'' and existing visual-frame prediction models briefly. Next, Section 3 introduces the network structure and methods in details. Section 4 shows the experimental results by making quantitative and qualitative evaluations of our methods compared with the baseline. Section 5 presents a brief discussion on the proposed method. Finally, Section 6 draws a conclusion and our thoughts about future studies.
\section{Related Work}
In order to better integrate predictive coding theory into neural networks, we need a detailed review of both aspects. In this section, the conceptual models of predictive coding and its related learning frameworks, as well as the state-of-the-art methods for visual-frame prediction from the perspective of machine learning will be reviewed.

\deleted{Conventional wisdom holds that the brain is like an input-output machine where any motor outputs as well as any decisions are merely results of the perception. However, the idea that ``the brain is predictive, not reactive" has been gradually accepted by cognitive scientists.} The predictive coding, which is a computational model of cognition, asserts   that our perception mostly comes from the brain's own internal inference model, combining sensory information with expectations. Those expectations can come from the current context, from internal model in the memory or as an ongoing prediction over time. As a theoretical ancestor, Helmholtz firstly proposed the concept of unconscious inference happened in the predictive brain \cite{von1867handbuch}. For example, an identical image can be perceived in different ways. Since the image formed on the retina does not change, perception must be the result of an unconscious process that deduces the cause of sensory information from top to down. Later in the 1940s', using empirical psychology study, Bruner demonstrated that perception is a result of the interaction between sensory stimuli (bottom-up as a recognition model) and conceptual knowledge (top-down as a generative model) \cite{bruner1947value}. Bar proposed a cognitive framework that the learned representation could be used in generating predictions, rather than passively ``waiting" to be activated by sensory input \cite{bar2007proactive}.  From the neuroscience perspective, Blom et al. also argued that predictions drive neural representations of visual events ahead of incoming sensory information \cite{blom2020predictions}, which suggests that the neural representations was driven by predictions generated by brain rather than the actual inputs.

\begin{figure}[]
	\centering{\includegraphics[width=4in]{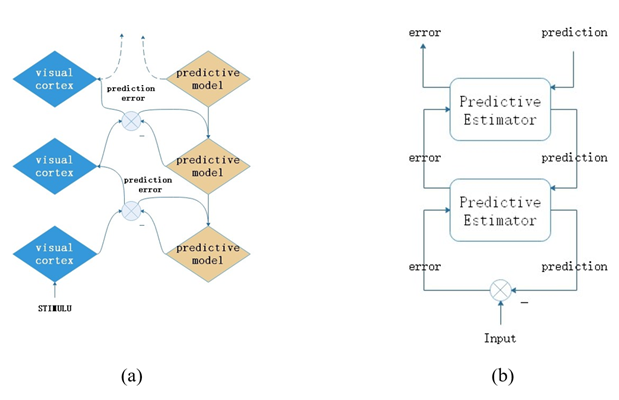}}
	\caption{A general framework of predictive coding. The visual cortex receives sensory inputs from outside world or signal errors from lower level to produce a local representation, which is then compared with the prediction made by predictive model. (b): Hierarchical network model for predictive coding proposed by Rao and Ballard \cite{rao1999predictive} }
	\label{fig:predictive coding}
\end{figure}

Depicting the predictive framework using a more rigorous expression, the term of ``predictive coding'' is imported from the field of signal processing. It is an algorithmic-based cognitive model aiming at the explanation of human cognition using the predictive framework. It has been applied in building computational models to explain different perceptual and neurobiological phenomena of the visual cortex \cite{watanabe2018illusory}. \deleted{ Similarly, the ``free energy'' principle \cite{friston2008hierarchical} depicts the process of prediction using hierarchical architectures. But in terms of the computational elements, the free energy principle utilizes the statistics of the signals, while some other models (e.g. \cite{rao1999predictive}) use error values themselves. In general, the predictive coding explains how the brain could learn the statistical regularities from the natural scenes as the experience. And such learned priors will be used to compare with sensory inputs to produce prediction error (i.e. mismatch). Such mismatching has been detected in neural activities, which implies the generative model is being updated in our predictive brain.}
\added{Specifically, it describes a simple hierarchical computational framework: neurons in higher level propagate predictions down, while neurons in lower level propagate prediction errors up. (as shown in Figure \ref{fig:predictive coding}) The entire model is updated through a combination of bottom-up and top-down information flows, so it does not rely solely on external information. Besides, the propagation of prediction errors constitutes effective feedback, allowing the model to perform self-supervised learning. The above characteristics make the predictive coding framework available and valuable to apply to the field of signal processing. For example, Whittington et al. proposed that a network developed in the predictive coding framework can efficiently perform supervised learning with simple local Hebbian plasticity. The activity of the prediction error node is similar to the error term in the backpropagation algorithm, so the weight change required by the backpropagation algorithm can be approximated by a simple Hebbian plasticity of connections in the prediction encoding network \cite{whittington2017approximation}.} 

\deleted{With the inspiration of predictive coding, there also have been a number of deep-learning based models aiming at solving practical problems with data-driving learning.} \added{In the field of visual-frame prediction, there is also a lot of work based on predictive coding.}   One of most successful applications is the PredNet model proposed by Lotter et al \cite{lotter2017deep}. It is a ConvLSTM-based model which stacked several ConvLSTMs vertically to generate a top-down propagation of prediction. On the other hand, the bottom-up propagation delivers the values of error. This model achieves the state-of-the-art performances in a few tasks such as video-frame prediction. Elsayed et al. \cite{elsayed2019reduced} implemented a novel ConvLSTM-based network called Reduced-Gate ConvLSTM which gives better performances. \added{However, although these works strictly follow the predictive coding style, the details are not well took into account. The predictive coding computational framework only roughly explain how the brain works, but some details, such as transmission delay, are ignored. The transmission delay has been discussed in the work of Hogendoorn et al. \cite{hogendoorn2019predictive} in detail. They pointed out that only when the concept of transmission delay is added, the predictive coding model can be regarded as a temporal prediction model. In addition, other neuroscientific phenomena, such as the different frequencies of oscillations in different levels of cortex, are equally important. Therefore, we designed a video prediction method with a comprehensive consideration of the different biological evidences mentioned above.}

Apart from the above methods, more predictive models were proposed building on the recent success of deep learning. The early state-of-the-art machine learning techniques are usually based on the encoder-decoder training. Using an end-to-end training manner, consecutive frames are used as inputs and outputs to train visual offsets or their semantic coherent. On the basis of the Encoder-Decoder network and LSTM, Villegas et al. proposed a novel method which decomposes the motion and content \cite{villegas2017decomposing} which encodes the local dynamics and the spatial layout separately, so as to simplify the task of prediction. \added{However, the motion referred to is simply obtained by subtracting $x_{t-1}$ from $x_t$. It describes changes at the pixel level only. }  Jin et al. \cite{jin2018varnet} also explored the inter-frame variations,  \added{which is similar to the MCNet. The innovation is that they} used  GDL (Gradient Difference Loss) regularization as loss function to sharpen predictions. In addition, Shi et al. also implemented the CNN-LSTM based model for precipitation nowcasting \cite{shi2015convolutional}. \added{Different from the previous two works, they embed convolutional neural networks directly into LSTM}, which makes better performance in capturing spatial-temporal correlations \added{and is also adopted in our network architecture.}  \deleted{This technique has been also used in a lot of video-frame prediction too.}

Besides \deleted{of temporal context training-based RNN or LSTM}, training in an adversarial fashion is another popular way, since the GAN (Generative Adversarial Network) shows excellent performance in image generation for prediction. For example, Aigner et al.\cite{aigner2018futuregan} proposed the FutureGAN based on the concept of PGGAN (Progressive Growing of GANs) in 2018. They extended this concept to the task of visual-frame prediction using a 3d convolutional encoder-decoder model to capture the spatial-temporal information. However, 3d convolution undoubtedly consumes more computation than other methods. Before PredNet, Lotter et al. have also proposed a GAN based model named Predictive Generative Network (PGN) which \cite{lotter2015unsupervised} training with a weighted MSE and adversarial loss for visual-frame prediction.

Summarized from the previous, \added{there are two main problems: 1)There is still room for improvement in network structure and training strategy. For instance, the Encoder-RNN-Decoder network only performs prediction in the high-level semantic space, resulting in most of the low-level details being ignored. 2)The computational cost is too large, consuming a lot of resources (especially during training). How to reduce the computational overhead through reasonable pruning is also important.} \deleted{most methods or  models have a common problem, that is, the computational cost is too large to consume a lot of resources (especially during training), and the prediction quality also needs to be improved.} \added{We have previously introduced the characteristics of predictive coding and the related theories, which provide an efficient and reliable theoretical computing framework. Therefore,} in order to reduce the consumption of resources and achieve sustainable artificial intelligence, we suggest combining the efficient cognitive framework and advanced data-driven machine learning methods to design an efficient predictive network model, which can not only improve predictive accuracy, but also reduce computational cost. Next, we will introduce our model in detail.
\section{Network Model and Methods}

\begin{figure}[]
	\centering{\includegraphics[width=4in]{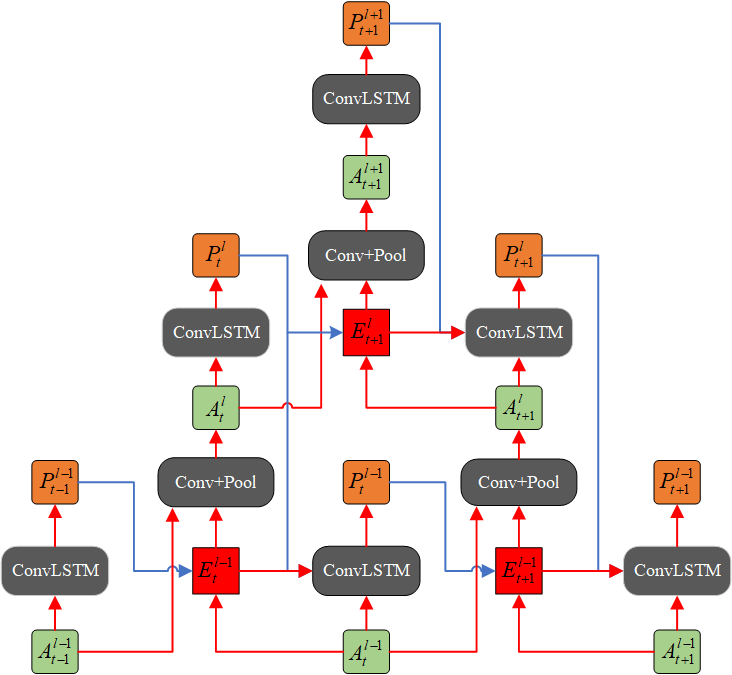}}
	\caption{Part of the PPNet. The green one denotes local sensory inputs of each layer while the orange one denotes local predictions, and the red one represents prediction errors. }
	\label{fig:PPNet}
\end{figure}

In this section, we will introduce the cognition-inspired model  which is specialised for visual-frame frame predictions. As its name (PPNet) refers to, its pyramid-like architecture is \replaced{beneficial}{benificial} to predict the visual frames as the neurons on the lower levels encode and predict the actual frames, and the neurons on-top encodes the scenarios which usually only change within a few visual frames (Figure \ref{fig:PPNet}). We will explain this idea in the next subsection. Then the detailed architecture as well as the algorithm will be introduced in the next few subsections. 

\subsection{Efficiency in Pyramid Architecture}
In this work, we mainly refer to the design concept of PredNet \cite{lotter2017deep} when building the network structure. As early as 2016, \replaced{Lotter}{lotter} et al. proposed such a typical predictive coding model which follows strictly the dual-way flow at every time-step and achieves outstanding performance. Nevertheless, the processing of information can be improved in at least two aspects. 

First, \replaced{according}{According} to predictive processing framework, at least two kinds of neurons are required: an internal representation neuron for generating predictions and an error calculation neuron for computing prediction errors. In the PredNet model, bottom-up inputs at each level are only served as targets of error calculation neurons for comparing with the top-down predictions to generate prediction errors, and the information carried in the upward propagation is only the prediction error itself. However,  we argue that it is necessary to use the past and present sensory information (represented here as video frames) as the inputs of representation neuron to generate predictions with higher accuracy. The formed memory can be formulated in a Bayesian framework, which is necessarily to be used to generate predictions. By learning such a Bayesian model, we can maximize the marginal likelihood or the entropy \cite{friston2010free}.

Second, as a cognitive-inspired model, we suggest such prediction and sensory input can be respectively implemented in at least two information streams in the hierarchical manner. This not only is inspired by our nervous system, but also is the way to integrate inputs from different network layers to get more spatiotemporal information which has been also widely used in the deep learning architecture, such as the ResNet, DenseNet and so on.

Based on the above assumptions, we propose and designed such a predictive model where the update rating of neurons on different levels can differ. \added{Alternatively, it can be also interpreted as a delay in information transmission. In general, it takes time for information to be transmitted from lower level to higher level, so there is a delay in transmission between different layers. However, neurons at the bottom layer do not passively wait for information transmitted from the top layer before making a prediction. The changes in biological synapses are determined only by the activity of presynaptic and postsynaptic neurons \cite{whittington2017approximation}. Therefore, in PPNet, once the prediction unit (ConvLSTM) receives sensory input (green), it will immediately combine with the prediction from higher level (if any) to make predictions. As what we mentioned in Sec.2, The delay of information transmission has been discussed in detail in the work of Hogendoorn et al \cite{hogendoorn2019predictive}. They argue that traditional predictive coding models such as the one first proposed by Rao and Ballard \cite{rao1999predictive} do not predict the future, but hierarchically predict what is happening. When the concept of transmission delay is added, the predictive coding model changes from hierarchical prediction to temporal prediction.} 

As a result, the PPNet could be regarded as an equivalent to the large-scale brain network (LSBN) where the higher cognitive function is conducted in the higher level of the deep learning network \deleted{, which is only updated while the error produced by the cognitive process exceeds certain threshold}. In the neuroscience evidence, such cognitive function which is processed in the PFC (prefrontal cortex) can be also used to predict the situated scenarios in our visual-frame prediction application for an agent. Therefore, our model is built considering the balance between biological evidence and the efficiency in computing.

\subsection{Network Architecture}
In this section, we will introduce our network model in detail. The architecture of our model is shown in Figure \ref{fig:PPNet}.
\added{
For the sake of reading and understanding, it is necessary to state the meaning of the symbols in the figure before making a detailed comparison and analysis:}

\begin{itemize}
	
	\item \added{\boldmath{$A_t^l$: }the green one, which represents the sensory input at level $l$ and time step $t$ }
	\item \added{\boldmath{$P_t^l$: }the orange one, which represents the prediction at level $l$ and time step $t$. Its prediction object is the sensory input at level $l$ and time step $t$ ($A_{t+1}^l$)}
	\item \added{\boldmath{$E_t^l$: }the red one, which represents the prediction error at level $l$ and time step $t$. It is calculated from previous prediction $P_{t-1}^l$ and current sensory input $A_t^l$.}

\end{itemize}

Inspired from PredNet, the PPNet also uses ConvLSTM as its basic components as they provide prediction flows with long-term dependency. Similarly, each layer of the network can be roughly divided into three parts: 

\begin{figure}[]
	\centering{\includegraphics[width=4.5in]{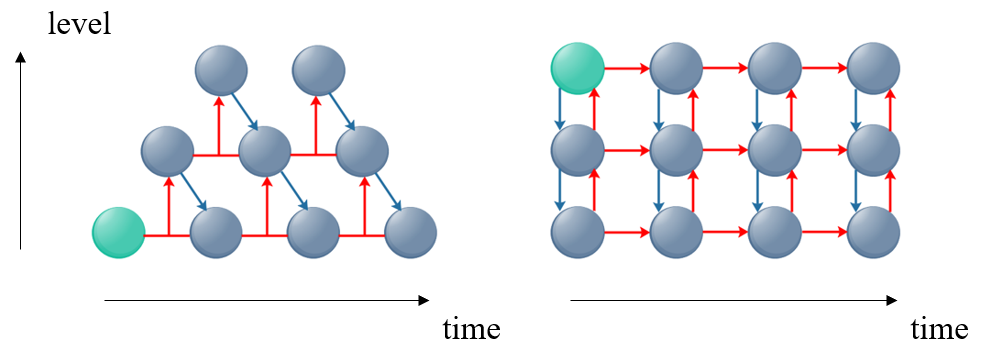}}
	\caption{The transmission of information in our model PPNet (left) and PredNet (right). The circle denotes an integration of the three parts mentioned above, and the green one is where the computation begins. The red arrows indicate the direction in which the only prediction errors (PredNet) or the combination of prediction errors and sensory inputs (PPNet) are propagated, while the blue arrows indicate the propagation of prediction from higher level.  }
	\label{fig:information}
\end{figure}

\begin{itemize}
	\item a predictive unit \deleted{(blue)}, which is made up of the recurrent convolutional network (ConvLSTM). It receives a sensory input \boldmath{$A_t^l$} \deleted{(or $x_t$ if in the first layer, the $x_t$ denotes the image at frame $t$ in the input visual sequence)} and a prediction \boldmath{$P_{t}^{l+1}$} from higher level (if any), to generate a local prediction \boldmath{$P_{t}^l$} of next time step.
	\item a generative unit \deleted{(yellow)}, which consist of a convolutional layer and a pooling layer. This unit is responsible for turning the local input $A_t^l$ \deleted{(or $x_t$)} as well as prediction error $E_{t+1}^l$ into the input $A_{t+1}^{l+1}$ of next level.
	\item an error representation layer \deleted{(red)}, which is split into separate rectified positive ($A_t^l-P_t^l$) and negative ($P_t^l-A_t^l$) error populations.
\end{itemize}

In order to process the prediction only when it is necessary, we show that the propagation of the dual-way can be done in a more efficient way. For a better understanding and comparison, a diagram (Figure \ref{fig:information}) is shown regarding the way of information propagation comparing our model and the PredNet model.

First, the computation of our model begins at the lowest layer after receiving the first sensory input, this is consistent with the design concept mentioned in section 3.1, which is different from PredNet which first starts at the top level by generating prediction without any prior information. Second, in our model, the bottom-up input of a higher-level unit comes from the combination of information from lower-level units of two time-steps.  Specifically, the current input \replaced{\boldmath{$A_{t}^l$}}{$A_{t-1}^l$}  is fed into internal representation neuron \added{(ConvLSTM)} to generate local prediction \boldmath{$P_{t}^l$}  at time step \boldmath{$t$}, which is then compared with next time step input \boldmath{$A_{t+1}^l$}to generate the prediction error \boldmath{$E_{t+1}^l$}. In other words, \boldmath{$A_{t+1}^l$} is not only a bottom-up sensory input for internal representation neuron at time step \boldmath{$t+1$}, but also the target of previous step \boldmath{$t$}, which is different from PredNet (the \boldmath{$A_{t+1}^l$} is just serve as a target at time-step \boldmath{$t+1$}).

Note with both prediction ($P_{t}^l$) and target ($A_{t+1}^l$) can PPNet generate prediction error for upward propagation. That is, at least two continuously sensory inputs $A_t^l$ and $A_{t+1}^l$ are required to generate prediction error for upward propagation, in which the former is served as an input to produce prediction while the latter is served as a target. As a result, the computation of neurons at different levels is not updated in a synchronized way at different levels, and the update frequency of neurons decreases as the network level increases, which is consistent with the biological evidence: deep neurons oscillate at a lower frequency \replaced{\cite{han2017rhythms}}{[1]}.  For this reason, the bottom-up input of top-level contains information for multiple time-steps at the bottom-level, which makes the PPNet has a stronger temporal correlation in structure, rather than relying solely on the temporal correlation of LSTM. In addition, it allows the PPNet to reduce the computational load by not having to update higher-level neurons.

\subsection{Training Loss and Adaptive Weight}

\begin{figure}[]
	\centering{\includegraphics[width=4in]{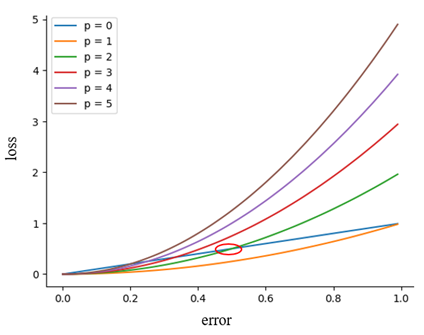}
		\caption{The relationship between hyper-parameter $p$ and the loss. Where $p = 0$ indicates that no weight is added, the original error is directly used as the loss value. The red circle marks the threshold between $p = 0$ and $p = 2$, indicating that when $p = 2$, more attention is paid to errors larger than the current threshold while less attention is paid to errors smaller than the threshold.}}
	\label{fig:error-loss}
\end{figure}

The training loss in our model is defined as the concatenation of positive and negative errors (Eq. \ref{eq:error}). where \replaced{$\hat{Y}$}{$P_t^l$} denotes prediction and \replaced{$Y$}{$A_t^l$} is target. \added{$ReLU$ denotes the ``rectified linear activation function", which is defined in Eq. \ref{eq:relu}. $concat$ means concatenating two multidimensional matrices together (for example, concatenating two matrices of dimension (b, c, h, w) into a matrix of (b, 2c, h, w)) } The Eq.\ref{eq:error} indicates the error population in the neurons incorporates both positive errors and negative errors \cite{rao1999predictive}. Furthermore, to sharpen the predictions, we introduce an adaptive weight into the loss function inspired by attention mechanism.

\begin{equation}
	E_t = concat[ReLU(\hat{Y}-Y), ReLU(Y-\hat{Y})] \label{eq:error}
\end{equation}

\begin{equation}
	f(x) =
	\begin{cases} 
		0,  &  x \leq 0 \\
		x, & x\textgreater \ 0
	\end{cases} \label{eq:relu}
\end{equation} 

At the beginning of the visual sequences, the error is usually quite large since it drives the top-down prediction to minimize the error. That is, the greater the prediction error, the stronger the brain response. We argue that the brain’s response can be seen as a weighting of the prediction error.  Based on this idea, we propose to add more weights to increase the contribution of prediction errors with higher values (for example, at the beginning of sequences). While the one with lower value, their contribution is reduced. A set of experiments raised by Kutas \& Hillyard \cite{kutas1980reading} have shown that, when prediction is seriously inconsistent with environment, the brain reacts more strongly.  Higher accuracy means less uncertainty, which is reflected in a higher gain on the relevant error units to do the update. In other words, the error units will become more adaptive to drive learning and plasticity if they are given an increasing weight. Therefore, we introduce a method of adaptive weights into our model, where a higher value of prediction error result in a higher weight.

\begin{equation}
	W_t = p E_t \label{eq:weight}
\end{equation}
\begin{equation}
	\mathcal{L}_{AW} = \sum_1^{T-1} \lambda_t W_t E_t \label{eq:loss}
\end{equation}

The adaptive weight for every time-step is calculated by directly multiplying the error itself by a coefficient (shown in Eq.\ref{eq:weight}). The $E_t$ denotes the prediction error at time step $t$, while the $p$ is a changeable hyper-parameter. So the training loss is defined in Eq.\ref{eq:loss}, where $T$ denotes the length of input sequences and $\lambda_t$ denotes the weighting factors by time. However, the error with a value less than $1/p$ will get smaller after being weighted. Figure 5 shows the relationship between $p$ and the loss. When the error is greater than the threshold (e.g., the intersection of the red circle), it will be enlarged. However, it will be reduced if it is less than the threshold. From an attention mechanism perspective, we pay more attention to errors larger than the threshold and pay less attention to errors smaller than the threshold. So the choice of threshold is extremely important. We will further explore the influence of hyper-parameter $p$ in the following experiments


\subsection{Algorithm}

In this section, we will introduce the algorithm to implement the above model based on the architecture and computation process mentioned in section 3.2. To better serve the following description, we reiterate the definition of each parameter as follow:

\begin{itemize}
	\item \boldmath{$E_t^l$}: the prediction error
	\item \boldmath{$H_t^l$}: the combination of hidden state $h_t^l$ and cell state $c_t^l$
	\item \boldmath{$A_t^l$}: the input as well as target of each layer
	\item \boldmath{$x_t$}: the image at frame $t$ in the input sequence
	\item \boldmath{$P_t^l$}: the prediction
	\item \boldmath{$T$}: the length of input sequence
\end{itemize}

\begin{equation}
	A_t^l =
	\begin{cases} 
		x_t,  & \text{if }l\text{ = 0} \\
		MaxPool(ReLU(Conv(E_t^{l-1}, A_{t-1}^{l-1}))), & \text{if }l\text{ \textgreater \ 0}
	\end{cases} \label{eq:3}
\end{equation} 
\begin{equation}
	H_{t}^l = ConvLSTM(A_t^l, H_{t-1}^l, upsample(P_{t}^{l+1})) \label{eq:4}
\end{equation}
\begin{equation}
	h_{t}^l, c_{t}^l = H_{t}^l\label{eq:5}
\end{equation}
\begin{equation}
	P_{t}^l = ReLU(Conv(h_{t}^l)) \label{eq:6}
\end{equation}
\begin{equation}
	E_{t+1}^l = [ReLU(P_{t}^l - A_{t+1}^l); ReLU(A_{t+1}^l - P_{t}^l)] \label{eq:7}
\end{equation}

\begin{algorithm}
	\caption{Calculation of the Pyramidal Predictive Network}
	\label{algorithm1}
	\LinesNumbered
	\KwIn{$A_t^0$ ${\leftarrow}$ $x_{1}, x_{2}, ..., x_{n}$ \\ \ \ \ \ \ \ \ \ \ \ \ \ $H_0^l$ ${\leftarrow}$ 0} 
	\KwOut{prediction of next frame $x_{n+1}$}
	\For{$t = 1 \ to \ T-1$}{	
		\For{$l = L \ to \ 0$}{
			\eIf{$A_t^l \ is \ None$}{
				$P_t^l = None$ \\
				$H_{t}^l = H_{t-1}^l$}{
				\eIf{$P_{t}^{l+1} \ is \ None$}{
					$H_{t}^l$ = $ConvLSTM(A_t^l, H_{t-1}^l)$
				}{$H_{t}^l$ = $ConvLSTM(A_t^l, H_{t-1}^l, upsample(P_{t}^{l+1}))$}
				$P_{t}^l$ = $ReLU(Conv(h_{t}^l))$}}
		\If{$t \textless T-1$}{
			\For{$l = 0 \ to \ L-1$}{
				\eIf{$P_{t}^l \ is \ None \ or \  A_{t+1}^l \ is \ None$}{
					$E_{t+1}^l$ = $E_t^l$ \\
					$A_{t+1}^{l+1} = None$
				}{$E_{t+1}^l$ = [$ReLU(P_{t}^l - A_{t+1}^l); ReLU(A_{t+1}^l - P_{t}^l$)] \\
					$A_{t+1}^{l+1}$ = $MaxPool(ReLU(Conv(E_{t+1}^l$, $A_t^l$)))}
			}
		}{}
	}
	
\end{algorithm}

The complete algorithms are listed in Eqs. \ref{eq:3} to \ref{eq:7}. The model is trained to minimized the training loss defined by Eq. \ref{eq:loss}, and our implementation is described in Algorithm \ref{algorithm1}. The information flows pass through two streams: \added{1)}A top-down propagation where the \replaced{hidden states $H_t^l$ of ConvLSTM is}{$H_{t+1}^l$ states are} updated and the local prediction $P_{t}^l$ is generated. \replaced{2)A bottom-up stream where}{and another stream delivers} the prediction errors $E_{t+1}^l$ \added{is calculated and propagated up to higher level along with the local input $A_t^l$.}\deleted{and the sensory inputs (as well as targets) $A_{t+1}^{l+1}$ of higher level are calculated.} Due to the pyramid design, the computation of our network updates at the lowest layer (i.e. layer $0$) at the first time-step. \added{However, for the convenience of programming, we refer to the programming method of PredNet, in which we perform the calculation of the top-down information flow first(line 2-11 in Algorithm \ref{algorithm1}), and then calculate the prediction error and update the sensory input of higher level (line 12-19 in Algorithm \ref{algorithm1}).} \deleted{but the algorithm will then update at higher level (i.e. $l=1$)  while the input $A_t^l$ appears at the time-step $t=1$. If no input at this time-step,} \added{Differently, if no sensory input $A_t^l$ at time-step $t$ and level $l$,} the calculation of this predictive unit is skipped without generating any predictions and the \added{hidden state of ConvLSTM} $H_{t}^l$ stays the same.

\section{Experiments}
In this section, several experiments are presented to show the performances of the PPNet using datasets for autonomous driving. We first introduce the features and pre-processing methods of three datasets: KTH, Caltech Pedestrian and KITTI, which are commonly used in the work of visual-frame prediction. Then the training detail and evaluations comparing PPNet and other state-of-the-art models will be presented in the following subsections.

\subsection{Dataset and pre-processing}
All the aforementioned datasets have to be processed into sequences before they can be used for training. In this part, we introduce the features of these datasets as well as the pre-processing methods.

\begin{itemize}
	\item \textbf{KTH}: The KTH dataset is an older dataset made in 2004 for human actions recognition. However, it is still very popular in the research of visual-frame prediction because of its simply scenario end events.
	\item \textbf{KITTI}: The KITTI dataset is one of the most widely-used datasets for autonomous driving. There are various processed data, but we directly download its raw images for training. Approx. 35K frames are used for training and 4.5K  for testing. The frames are center-cropped and resized to $128 \times 160$ pixels in the same way as PredNet. Compared to the other two datasets, its variations of interframe are greater.
	\item \textbf{Caltech Pedestrian}: It was originally designed for pedestrian detection, which is also suitable for the work of visual-frame prediction. The frames are directly resized to $128 \times 160$ pixels which is the same as KITTI. The variations of its interframe are much smaller than KITTI, which might result in the model learns a repetition instead of prediction.
\end{itemize}

\subsection{Training Setting}
We implemented the PPNet using PyTorch platform and trained it on a Geforce RTX 3070 GPU. The length of input sequence is set to 10 and the number of layers in the network is set to 6. Other hyper-parameters are shown in Table \ref{table:1}. Influenced by the initialization, the time-weight $\lambda_t$ of the prediction error generated at the first time step is set to 0.5, while the rest are set to 1.

\begin{table}[t]
	\caption{Hyper-parameters for training, including training epoch, learning rate, hyper-parameter $p$ and $\lambda_t$ defined in Section 3.3.}
	\begin{center}

		\begin{tabular}{lccc}
			\hline
			\multirow{2}{*}{hyper-parameters} & \multicolumn{3}{c}{Datasets}                   \\ 
			& KITTI       & Caltech Pedestrian      & KTH     \\ \hline
			
			epoch                     & 300         & 200                    & 200     \\ 
			learning rate                     & \multicolumn{3}{c}{0.0002}                     \\ 
			\textit{P}                        & $10^4$      & $10^4$                 &$10^3$   \\ 
			
			$\lambda_t$                                & \multicolumn{3}{c}{$\lambda_t =
				\begin{cases} 
					0.5,  & \text{if }t\text{ = 0} \\
					1, & \text{if }t\text{ \textgreater \ 0}
				\end{cases}$}             \\  \hline
		\end{tabular}
	\end{center}
	\label{table:1}
\end{table}

\begin{figure}[t]
	\centering{\includegraphics[width=4.5in]{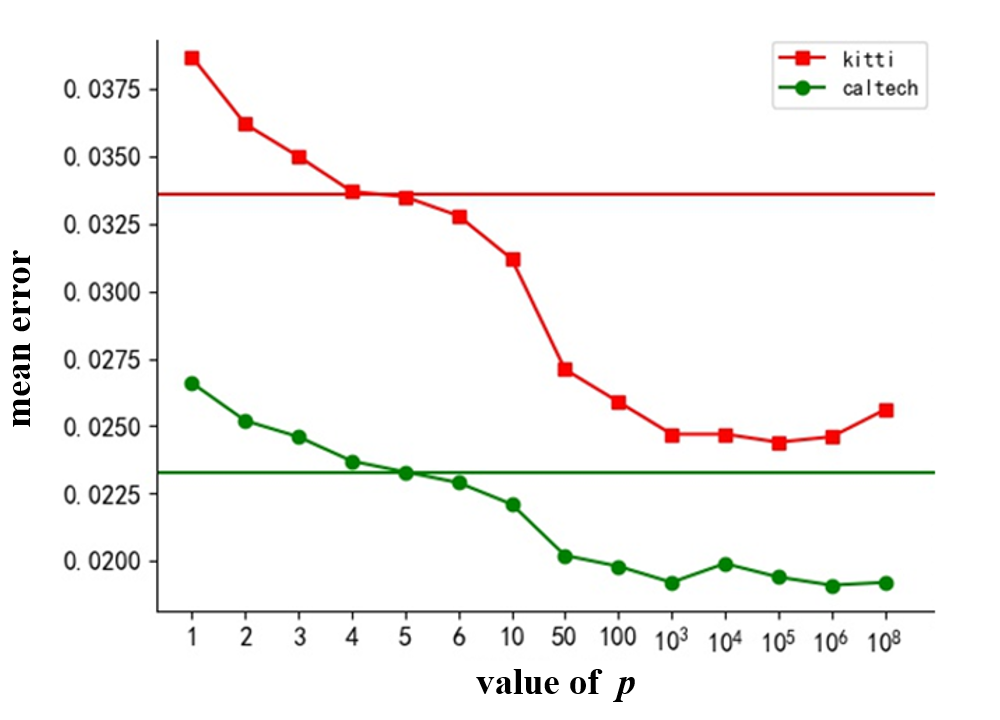}}
	\caption{Experiment results with KITTI dataset (red) and Caltech Pedestrian Dataset (green) using different value of $p$. The horizontal lines indicate the results without adding any weight.  }
	\label{fig:p and error}
\end{figure}

In order to pick up a suitable value for the hyper-parameter p proposed previously, we have performed two sets of experiments using part of the KITTI dataset and Caltech Pedestrian dataset to explore its influence, the results are shown in Figure \ref{fig:p and error}. The horizontal lines indicate the results without adding any weight. According to Eq. \ref{eq:weight} and Eq. \ref{eq:loss}, when the value of p is set to 1, the loss function will be equivalent to the mean square error loss. Obviously, the method of dividing the error into positive error and negative error is indeed beneficial Better results can be observed while the value of p is greater than 5 (or 6) compared to the one without any weight. The training loss (mean errors) is getting smaller with the increasing of p, and we got almost the best result while it is close to ${10}^3$. However, it might result in an opposite performance if keeping increasing its value. So we chose the value around ${10}^3$ in the subsequent experiments.

\subsection{Evaluation Results}
In this section, we use SSIM \cite{wang2004image}, PSNR \cite{hore2010image} and LPIPS \cite{zhang2018unreasonable} for quantitative evaluation. SSIM is an early measure of image similarity, which compares two images from the perspective of brightness, contrast, and structure. PSNR is also a metric for evaluating image quality. It measures the degree of image distortion by calculating the ratio of the maximum signal to background noise. However, the above two evaluation indicators have the same problem: the results may not match the evaluation of the human visual system \cite{vcadik2012new}. To solve this problem, Zhang et al. proposed the LPIPS metric to try to simulate the evaluation of human visual system. Higher values indicate better results for SSIM and PSNR, while lower values indicate better results for LPIPS.
~\\

\begin{table}[h]
	\centering
	
	\caption{The quantitative evaluation results on the KTH dataset. The results are averaged from future 10 time steps ($10\rightarrow20$) and 30 time steps ($10\rightarrow40$) respectively.}
	\label{Tab:KTH}
	\resizebox{\linewidth}{!}
	{
		\begin{tabular}{lcccccc}
			\hline
			\multirow{2}{*}{Methods} & \multicolumn{3}{c}{10 $\rightarrow$ 20} & \multicolumn{3}{c}{10 $\rightarrow$ 40} \\
			& SSIM $\uparrow$  & PSNR $\uparrow$  & LPIPS $\downarrow$ 
			& SSIM $\uparrow$  & PSNR $\uparrow$  & LPIPS $\downarrow$  \\
			\hline
			MCNet  \cite{villegas2017decomposing}                  & 0.804  & 25.95  & -     & 0.73   & 23.89  & -      \\
			fRNN   \cite{oliu2018folded}                  & 0.771  & 26.12  & -     & 0.678  & 23.77  & -      \\
			PredRNN  \cite{wang2017predrnn}                & 0.839  & 27.55  & -     & 0.703  & 24.16  & -      \\
			PredRNN++   \cite{wang2018predrnn++}             & 0.865  & 28.47  & 22.89     & 0.741  & 25.21  & 27.90      \\
			VarNet  \cite{jin2018varnet}                 & 0.843  & 28.48  & -     & 0.739  & 25.37  & -      \\
			
			E3D-LSTM \cite{wang2018eidetic}                & 0.879  & 29.31  & 29.84     & 0.810  & 27.24  & 32.88     \\
			Conv-TT-LSTM \cite{su2020convolutional}               & \textbf{0.907}  & 28.36  & 13.34 & \textbf{0.882}  & 26.11   & 19.12  \\
			LMC-Memory \cite{lee2021video}  & 0.894  & 28.61 & 13.33 &0.879  & 27.50  & \textbf{15.98} \\
			\hline
			Ours      &  0.886      &  \textbf{31.02}     & \textbf{ 13.12 }    &  0.821      &    \textbf{28.37}    &   23.19    \\
			\hline
		\end{tabular}
	}
	
\end{table}

\begin{figure}[h]
	\centering{\includegraphics[width=5.5in]{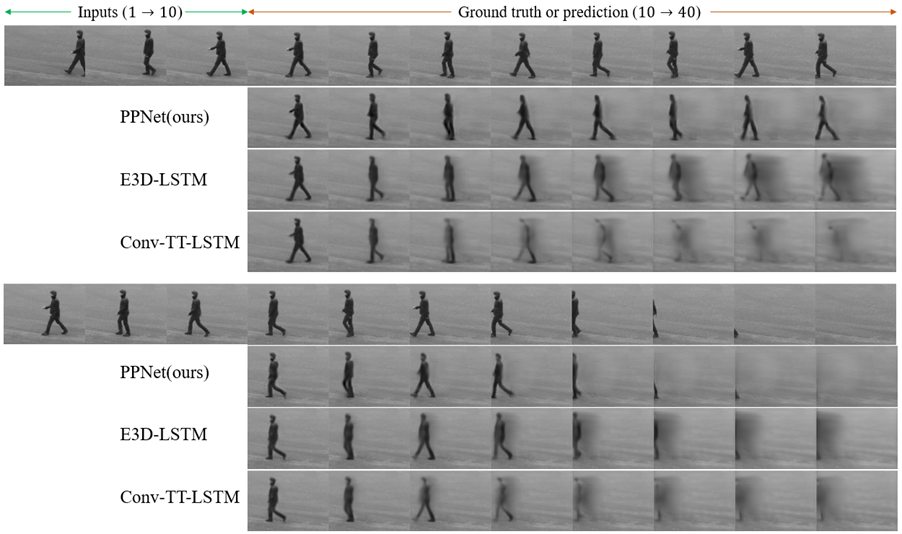}}
	\caption{The visual presentation of predicted frames on the KTH dataset. We take 10 frames as input and predict the next 30 frames}
	\label{fig:KTH}
\end{figure}

\textbf{Results on KTH dataset} \quad Table \replaced{\ref{Tab:KTH}}{1}  shows the quantitative evaluation results with the state-of-the-art methods on the KTH dataset. Similar to previous work, we made calculations on the average results over the future 10 frames ($10\rightarrow20$) and 30 frames ($10\rightarrow30$) respectively, with 10 input frames. Our method does achieve better or comparable results compared with the state-of-the-art works in terms of accuracy assessment. However, in the field of video prediction task, pure quantitative evaluation seems to be weak sometimes. Therefore, we also visualized the predicted results. Figure \ref{fig:KTH} shows the predicted examples of our method and other proposed methods. Obviously, our method also achieves good results from the perspective of human visual system evaluation, while the Conv-TT-LSTM \cite{su2020convolutional}, which has acquired outstanding performance in quantitative evaluation, performs poorly from the perspective of visual presentation (Actually, it also performed poorly in work \cite{lee2021video}). This is a common problem in video prediction tasks. There is not an accurate and uncontroversial evaluation metric like image classification or semantic segmentation. As a result, we need to combine the quantitative evaluation and qualitative evaluation to make a better comparison.

\begin{table}[H]
	\centering
	
	\caption{The quantitative evaluation results on the Caltech and KITTI datasets respectively. The results are averaged from future 5 time steps ($10\rightarrow15$).}
	\label{Tab:Caltech-KITTI}
	\resizebox{\linewidth}{!}
	{
		\begin{tabular}{lcccccc}
			\hline
			\multirow{2}{*}{Methods} & \multicolumn{3}{c}{Caltech 10 $\rightarrow$ 15} & \multicolumn{3}{c}{KITTI 10 $\rightarrow$ 15} \\
			& SSIM $\uparrow$  & PSNR $\uparrow$  & LPIPS $\downarrow$ 
			& SSIM $\uparrow$  & PSNR $\uparrow$  & LPIPS $\downarrow$  \\
			\hline
			MCNet  \cite{villegas2017decomposing}                  & 0.705  & -  & 37.34     & 0.555   & -  & 37.39     \\
			
			PredNet  \cite{lotter2017deep}                 & 0.752  & -  & 36.03     & 0.475  & -  & 62.95     \\
			
			Voxel Flow \cite{liu2017video}               & 0.711  & -  & 28.79    & 0.426  & -  & 41.59     \\
			Vid2vid \cite{wang2018video}               & 0.751  & -  & 20.14 & -  & -   & -  \\
			FVSOMP \cite{wu2020future}  & 0.756  & - & 16.50 &0.608  & - & 30.49 \\
			\hline
			Ours      &  \textbf{0.812}      &  \textbf{21.3}     & \textbf{ 14.83 }    &  \textbf{0.617}      &    \textbf{18.24}    &   31.07   \\
			\hline
		\end{tabular}
	}
	
\end{table}

\begin{figure}[H]
	\centering{\includegraphics[width=5.5in]{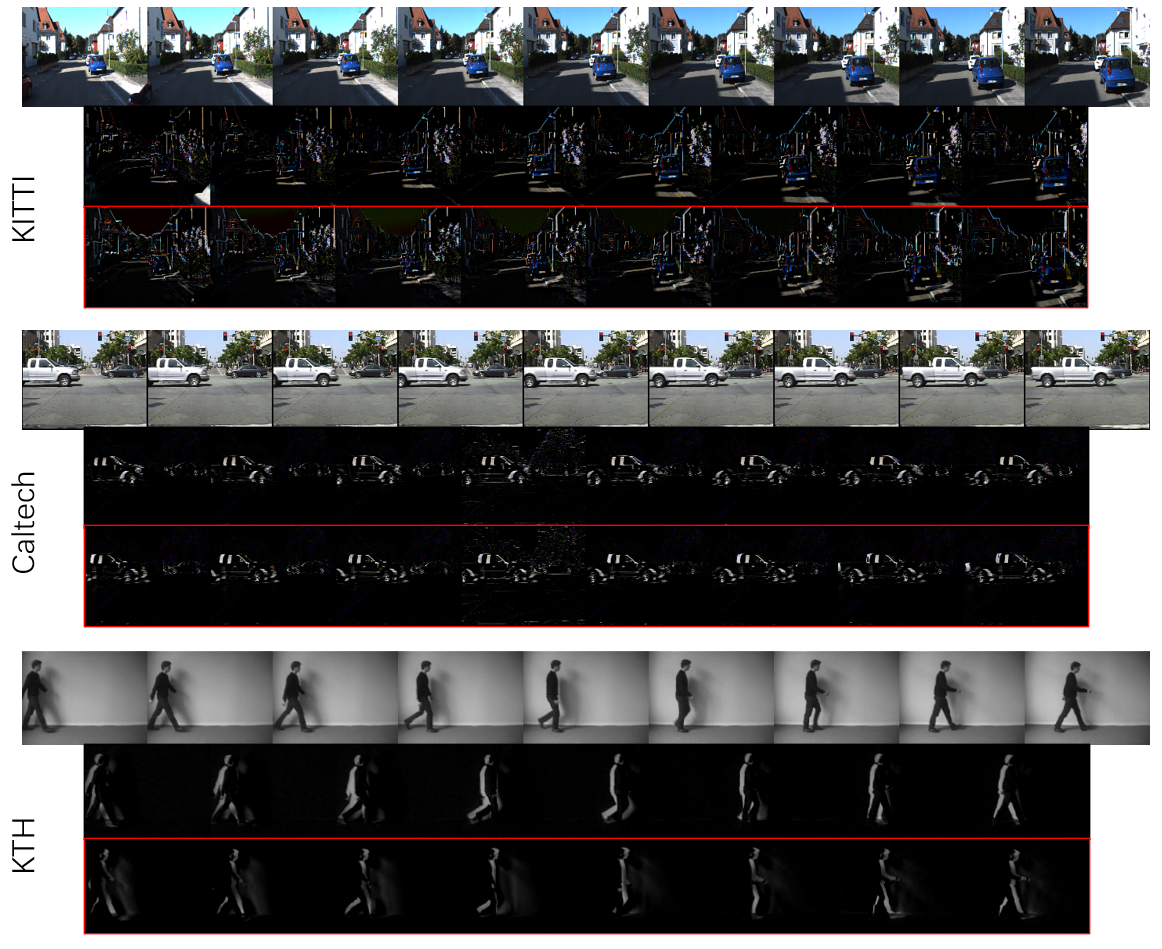}}
	\caption{Visualization of variations between frames in each dataset. In each group, the first row indicates the raw frames, the second row indicates positive variations and the last row denotes negative variations. }
	\label{fig:variation}
\end{figure}

\textbf{Results on Caltech and KITTI dataset} \quad We also validate our methods on Caltech and KITTI datasets, which have more complex scenarios and events. Table \ref{Tab:Caltech-KITTI} shows the quantitative evaluation results. Obviously, even though we only count the predicted frames of future 5 time-steps, the results are still much worse than the performance on KTH. In fact, it has to do with how complex and varied the scene is. The more complex the scene and the greater the variation, the more difficult it is to predict. As shown in Figure \ref{fig:variation}, we visualized the inter-frames variation of the three datasets separately. The Catech has a similar level of sophistication as KITTI, but KITTI is more variable than Caltech and therefore the methods perform worse on KITTI. Prediction in complex scenes is also an urgent problem to be solved in current video prediction tasks.
~\\

\begin{table}[H]
	\centering
	\caption{Evaluation of next frame prediction on each dataset. We make comparison in terms of prediction accuracy and computational overhead.}
	\label{Tab:PPNet-PredNet}
	\resizebox{\linewidth}{!}
	{
		\begin{tabular}{lcccccc}
			\hline
			\multirow{2}{*}{Metrics} & \multicolumn{2}{c}{KTH}  & \multicolumn{2}{c}{Caltech} & \multicolumn{2}{c}{KITTI} \\
			& Ours           & PredNet & Ours             & PredNet  & Ours            & PredNet \\ \hline
			SSIM $\uparrow$                    & \textbf{0.945} & 0.934   & \textbf{0.919}   & 0.887    & \textbf{0.787}  & 0.642   \\
			PSNR $\uparrow$                    & \textbf{36.47} & 33.31   & \textbf{28.44}   & 23.56    & \textbf{21.96}  & 16.58   \\
			LPIPS  $\downarrow$                  & \textbf{8.03}  & 8.92    & \textbf{7.35}    & 14.65    & \textbf{21.49}  & 38.51   \\
			Time/ms $\downarrow$                 & \textbf{27.6}  & 52.2    & \textbf{37.0}    & 71.4     & \textbf{37.2}   & 71.5   \\ \hline
		\end{tabular}
	}
\end{table}

\begin{figure}[H]
	\centering{\includegraphics[width=5.5in]{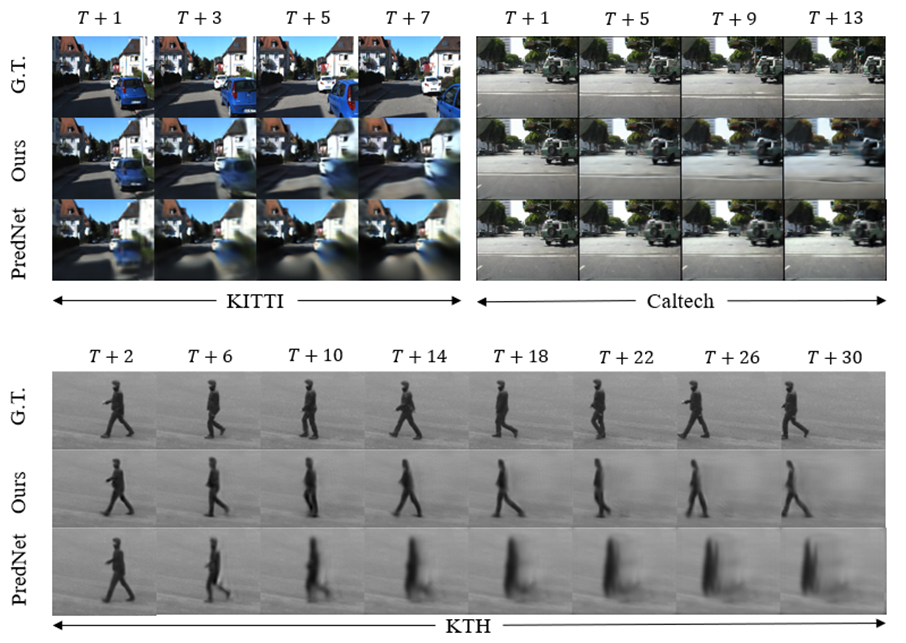}}
	\caption{The visual presentation of predicted frames on the KITTI, Caltech and KTH datasets respectively. }
	\label{fig:PPNet-PredNet}
\end{figure}

\textbf{Comparison with PredNet} \quad \added{As what we mentioned above, the PredNet strictly follows the computational style of  traditional predictive coding framework, and the network structure of PPNet and PredNet is similar (for example, both use ConvLSTM as the backbone).} \deleted{Compared to the other methods, the architecture of PPNet  is most similar to the PredNet mentioned above} (the PredNet model is redrawn in the same way as our model in Appendix \added{A}). Therefore, it is easier to set the same parameters such as network depth and width to retrain the PredNet to make a fair and clear comparison, which can be considered as ablation study, to highlight the rationality and superiority of our model. Table \ref{Tab:PPNet-PredNet} shows the performances of next frame prediction on KTH, Caltech and KITTI respectively. Obviously, our method is superior to PredNet in both prediction accuracy and computational overhead. The pyramid style is effective. By reducing the oscillation frequency, higher-level neurons can not only obtain longer-term information, but also reduce the computational cost. 

Figure \ref{fig:PPNet-PredNet} visualizes the long-term predictions on each dataset with different predicted time steps respectively. In general, our results are better than those of PredNet. First, it can be seen from the figure that the inter-frame variation of the KITTI dataset is much larger than that of the other two, both PPNet and PredNet made the fuzzy predictions. However, the PPNet can still make better predictions in the first few steps while PredNet makes blurry predictions and then reproduces them only. This kind of replication is more obvious when using the Caltech dataset for evaluation. Though generating clearer frames compared to our method, the PredNet is just reproducing previous frames instead of making predictions. On the contrary, the PPNet is still able to capture the motion information in the input sequences and make authentic predictions. PredNet captures the motion information on the KTH dataset finally, but it learned only the person’s direction and an approximated speed, while other subtle movements, such as the actions of the person’s arm and leg, are lost. 
~\\

In summary, we have presented several experiment results to show a remarkable performance of our method, \replaced{which is superior}{which superior} to PredNet in terms of prediction accuracy, computational cost and visual presentation. In addition, we can also get results equal to or better than the other state-of-the-art methods, indicating the superiority of our method.

\section{Discussion}

\subsection{Propagation of Weighted Error}
Additional experiments were performed to explore the influence of the propagation of prediction errors. According to what is mentioned above, the prediction errors will be propagated upward for higher level. Here comes the question: which errors should be pass up, the original errors or the weighted errors? It is necessary to indicate that the result shown in Figure \ref{fig:p and error} is in which the original errors were transmitted upward and the weighted errors were only propagated backward. We did get a worse result while we propagated the weighted errors both upward and backward after being normalized (Table \ref{Tab:p-error}). As Corlett \cite{corlett2009drugs} and Fletcher et al. \cite{fletcher2009perceiving} have speculated: errors might be ``false" after being weighted, it will make profound corrections to our model of the world if waves of persistent and highly weighted ``false errors" were propagated upward. Using the adaptive weights proposed in Section 3.3, we provide a possible proof for the assumption from the perspective of artificial neural network.

\begin{figure}[H]
	\centering{\includegraphics[width=5.5in]{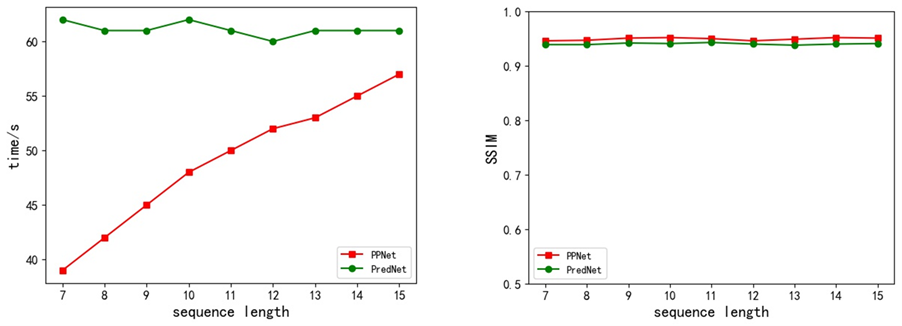}}
	\caption{Evaluation on KTH dataset using input sequence with different length. The left figure shows the time required for each training epoch, and the right one shows the prediction accuracy. }
	\label{fig:lenth-time}
\end{figure}

\begin{table}[t]
	\begin{center}
		\caption{The mean errors (ME) obtained using different ways of propagation.}
		\resizebox{\linewidth}{!}{
			\begin{tabular}{lcccccc}
				\hline
				\multicolumn{2}{l}{value of   $p$}                & 5      & 10     & 100    & 1000   & 10000  \\ \hline
				\multirow{2}{*}{backward}             & kitti   & 0.0335 & 0.0312 & 0.0259 & 0.0247 & 0.0247 \\ 
				& caltech & 0.0233 & 0.0221 & 0.0198 & 0.0192 & 0.0199 \\ 
				\multirow{2}{*}{backward and upward} & kitti   & 0.0450 & /      & /      & 0.0463 & 0.0460 \\ 
				& caltech & /      & 0.0303 & 0.0311 & 0.0316 & /      \\ \hline
		\end{tabular}}
	\label{Tab:p-error}
	\end{center}
\end{table}

\subsection{The Efficiency of Pyramid-like Architecture}
A set of priors is often already active on a higher level of cognitive hierarchy, poised to impact the processing of new sensory inputs without further delay while the context information have been in place. Similarly, there is a delay in the upward flow of information at the beginning, but it will disappear once the information reaches the highest level in our model, which might result in a trivial reduction of computational cost while the input sequence is long enough. However, longer sequences are not required. LSTM networks may capture spurious long-term dependencies that may have been present in the training data, hence learning inadequate causal models \cite{pearl2009causality}. Additionally, we have performed a set of experiments on both PPNet and PredNet by processing the same data into sequences with different length to prove our point (Note that the total number of video frames is constant). As shown in Figure \ref{fig:lenth-time}, the length of input sequence has little effects on the prediction accuracy, but less time was required using a shorter sequence in our proposed PPNet. Therefore, we can process the data into shorter sequences during training, to reduce the consumption of resources and achieve sustainable artificial intelligence.

\section{Conclusions}

In this paper, we have demonstrated a pyramidal predictive network for visual-frame prediction based on the predictive coding concept, with much efficient computational manner. This model encodes information at various temporal and spatial scale, with a up-down propagation of prediction and a bottom-up propagation of the combination of sensory input and prediction error. It has a stronger temporal correlation in structure and uses less computation cost. We analyzed the rationality of the model in detail from the perspectives of predictive processing and machine learning. Importantly, this proposed model achieve a remarkable performance compared to state-of-the-art models according to the experimental results. 

Nevertheless, there is still room for improvement for the proposed model. In the long-term forecasting process, the false ``prediction errors" may cause the model to average the possible future into a single, fuzzy forecast, which is an urgent problem exists in most predictive model. In addition, prediction on directly predicting natural visual frames is still a challenging task due to the curse of dimensionality. Therefore, in the future, we are going to reduce the prediction space to high-level representations, such as semantic and instance segmentation, and depth space, to simplify the prediction task, which will make the intelligent robots easier to predict and perform advanced actions.

\vspace{6pt} 



\authorcontributions{Experiment, C.L.; writing original draft preparation, C.L.; review and editing, J.Z. and H.L. }

\funding{This work was supported in part by the Key-Area Research and Development Program of Guangdong Province under Grant: 2019B090912001, and in part by the PolyU Start-up Grant: ZVUY-P0035417.}

\dataavailability{Code is available at \url{https://github.com/Ling-CF/PPNet}.} 

\acknowledgments{The authors would like to thank the reviewing from editors and reviewers.}

\conflictsofinterest{The authors have no conflicts of interest to declare.} 



\appendixtitles{no} 
\appendixstart
\appendix
\begin{figure}[h]
	\centering{\includegraphics[width=5.5in]{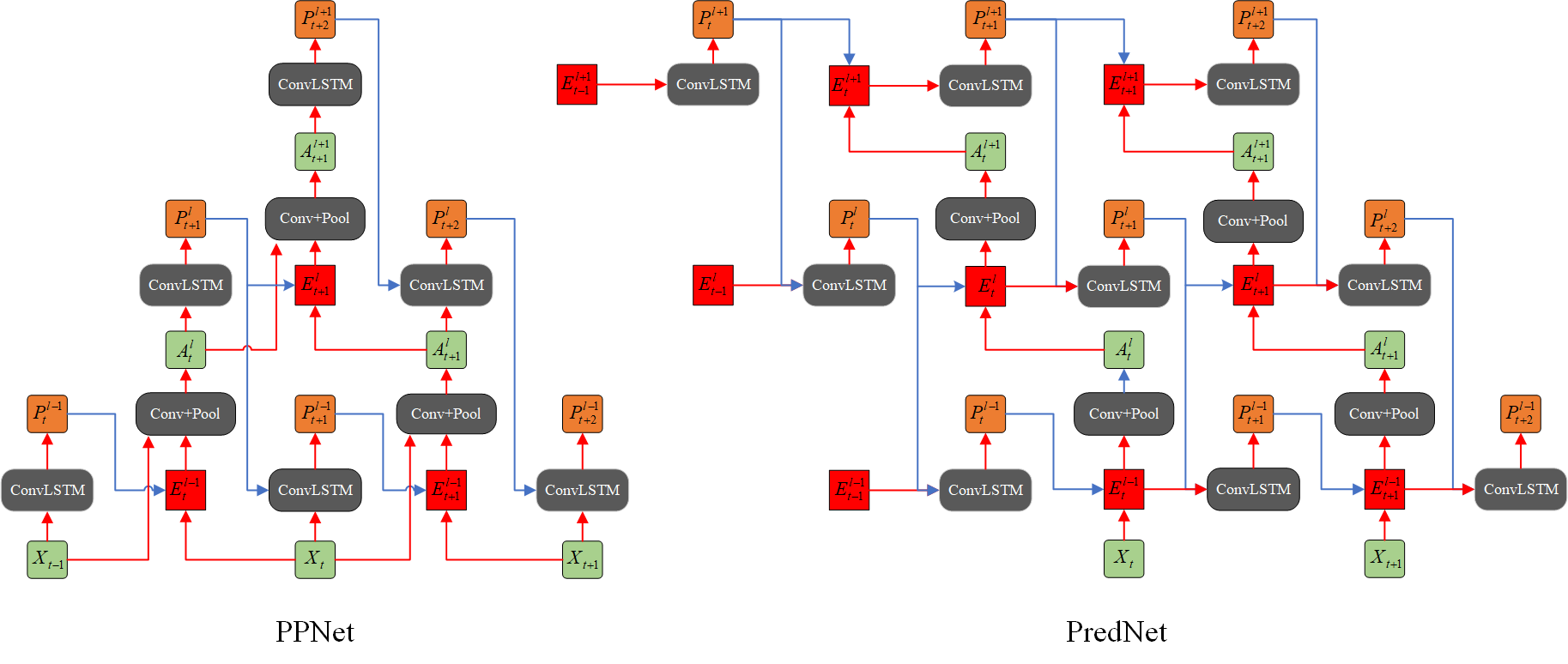}}
	\caption{The network structures of PPNet and PredNet, where the PredNet is redrawn in the same way as PPNet, for a better comparison. }
	\label{fig:PPN-PredNet-Network}
\end{figure}
\section[\appendixname~\thesection]{}
We provide a clear comparison of our model and PredNet here, to further illustrate the differences between the two models. \replaced{Figure \ref{fig:PPN-PredNet-Network}}{Figure 11}  shows the architecture of the two models, where PredNet is redrawn in the same way as PPNet. As shown in the figure, the biggest difference is that the update frequency of neurons decreases as the network level increases (ConvLSTM, etc.), while in PredNet, neurons of all levels are calculated and updated at each time step. Therefore, in our model, higher-level neurons can receive information from longer time series with less computational overhead, and this advantage is more pronounced as more network layers are stacked. 

In addition, the computational concepts of the two models are different. PredNet considers that the prediction is generated by the internal model first, so in this model, the prediction is first made at the top layer, then passed down to the lowest layer, and finally compared with the sensory input (the green one) to get the prediction error which is then pass up to higher level. On the contrary, we believe that there should be sensory input before prediction (discussed in Section 3.1: Efficiency in Pyramid Architecture), so in our model, the lowest level neurons first receive sensory input and make predictions, and the information is passed up only after the prediction error is obtained by comparing the current prediction with  sensory input of next time step. Besides, the information we transmit upward includes not only predictive error but also sensory input information, the reasons have also been explained in Section 3. The above is the main difference between our model and PredNet in terms of network structure.

~\\

\begin{adjustwidth}{-\extralength}{0cm}

\reftitle{References}



\bibliographystyle{mdpi}
\bibliography{ref1}

\begin{thebibliography}{999}

\bibitem[Morris and Trivedi(2008)]{morris2008learning}
Morris, B.T.; Trivedi, M.M.
\newblock Learning, modeling, and classification of vehicle track patterns from
  live video.
\newblock {\em IEEE Transactions on Intelligent Transportation Systems} {\bf
  2008}, {\em 9},~425--437.

\bibitem[Kitani \em{et~al.}(2012)Kitani, Ziebart, Bagnell, and
  Hebert]{kitani2012activity}
Kitani, K.M.; Ziebart, B.D.; Bagnell, J.A.; Hebert, M.
\newblock Activity forecasting.
\newblock In Proceedings of the European Conference on Computer Vision.
  Springer,  2012, pp. 201--214.

\bibitem[Bhattacharyya \em{et~al.}(2018)Bhattacharyya, Fritz, and
  Schiele]{bhattacharyya2018long}
Bhattacharyya, A.; Fritz, M.; Schiele, B.
\newblock Long-term on-board prediction of people in traffic scenes under
  uncertainty.
\newblock In Proceedings of the Proceedings of the IEEE Conference on Computer
  Vision and Pattern Recognition,  2018, pp. 4194--4202.

\bibitem[Shi \em{et~al.}(2017)Shi, Gao, Lausen, Wang, Yeung, Wong, and
  Woo]{shi2017deep}
Shi, X.; Gao, Z.; Lausen, L.; Wang, H.; Yeung, D.Y.; Wong, W.k.; Woo, W.c.
\newblock Deep learning for precipitation nowcasting: A benchmark and a new
  model.
\newblock {\em Advances in neural information processing systems} {\bf 2017},
  {\em 30}.

\bibitem[Softky(1996)]{softky1996unsupervised}
Softky, W.R.
\newblock Unsupervised pixel-prediction.
\newblock In Proceedings of the Advances in neural information processing
  Systems,  1996, pp. 809--815.

\bibitem[Deco and Sch{\"u}rmann(2001)]{deco2001predictive}
Deco, G.; Sch{\"u}rmann, B.
\newblock Predictive coding in the visual cortex by a recurrent network with
  gabor receptive fields.
\newblock {\em Neural processing letters} {\bf 2001}, {\em 14},~107--114.

\bibitem[Von~Helmholtz(1867)]{von1867handbuch}
Von~Helmholtz, H.
\newblock {\em Handbuch der physiologischen Optik: mit 213 in den Text
  eingedruckten Holzschnitten und 11 Tafeln}; Vol.~9, Voss,  1867.

\bibitem[Bruner and Goodman(1947)]{bruner1947value}
Bruner, J.S.; Goodman, C.C.
\newblock Value and need as organizing factors in perception.
\newblock {\em The journal of abnormal and social psychology} {\bf 1947}, {\em
  42},~33.

\bibitem[Bar(2007)]{bar2007proactive}
Bar, M.
\newblock The proactive brain: using analogies and associations to generate
  predictions.
\newblock {\em Trends in cognitive sciences} {\bf 2007}, {\em 11},~280--289.

\bibitem[Blom \em{et~al.}(2020)Blom, Feuerriegel, Johnson, Bode, and
  Hogendoorn]{blom2020predictions}
Blom, T.; Feuerriegel, D.; Johnson, P.; Bode, S.; Hogendoorn, H.
\newblock Predictions drive neural representations of visual events ahead of
  incoming sensory information.
\newblock {\em Proceedings of the National Academy of Sciences} {\bf 2020},
  {\em 117},~7510--7515.

\bibitem[Watanabe \em{et~al.}(2018)Watanabe, Kitaoka, Sakamoto, Yasugi, and
  Tanaka]{watanabe2018illusory}
Watanabe, E.; Kitaoka, A.; Sakamoto, K.; Yasugi, M.; Tanaka, K.
\newblock Illusory motion reproduced by deep neural networks trained for
  prediction.
\newblock {\em Frontiers in psychology} {\bf 2018}, {\em 9},~345.

\bibitem[Whittington and Bogacz(2017)]{whittington2017approximation}
Whittington, J.C.; Bogacz, R.
\newblock An approximation of the error backpropagation algorithm in a
  predictive coding network with local hebbian synaptic plasticity.
\newblock {\em Neural computation} {\bf 2017}, {\em 29},~1229--1262.

\bibitem[Rao and Ballard(1999)]{rao1999predictive}
Rao, R.P.; Ballard, D.H.
\newblock Predictive coding in the visual cortex: a functional interpretation
  of some extra-classical receptive-field effects.
\newblock {\em Nature neuroscience} {\bf 1999}, {\em 2},~79--87.

\bibitem[Lotter \em{et~al.}(2017)Lotter, Kreiman, and Cox]{lotter2017deep}
Lotter, W.; Kreiman, G.; Cox, D.
\newblock Deep predictive coding networks for video prediction and unsupervised
  learning.
\newblock {\em International Conference on Learning Representations} {\bf
  2017}.

\bibitem[Elsayed \em{et~al.}(2019)Elsayed, Maida, and
  Bayoumi]{elsayed2019reduced}
Elsayed, N.; Maida, A.S.; Bayoumi, M.
\newblock Reduced-Gate Convolutional LSTM Architecture for Next-Frame Video
  Prediction Using Predictive Coding.
\newblock In Proceedings of the 2019 International Joint Conference on Neural
  Networks (IJCNN). IEEE,  2019, pp. 1--9.

\bibitem[Hogendoorn and Burkitt(2019)]{hogendoorn2019predictive}
Hogendoorn, H.; Burkitt, A.N.
\newblock Predictive coding with neural transmission delays: a real-time
  temporal alignment hypothesis.
\newblock {\em Eneuro} {\bf 2019}, {\em 6}.

\bibitem[Villegas \em{et~al.}(2017)Villegas, Yang, Hong, Lin, and
  Lee]{villegas2017decomposing}
Villegas, R.; Yang, J.; Hong, S.; Lin, X.; Lee, H.
\newblock Decomposing motion and content for natural video sequence prediction.
\newblock In Proceedings of the International Conference on Learning
  Representations,  2017.

\bibitem[Jin \em{et~al.}(2018)Jin, Hu, Zeng, Tang, Liu, and Ye]{jin2018varnet}
Jin, B.; Hu, Y.; Zeng, Y.; Tang, Q.; Liu, S.; Ye, J.
\newblock Varnet: Exploring variations for unsupervised video prediction.
\newblock In Proceedings of the 2018 IEEE/RSJ International Conference on
  Intelligent Robots and Systems (IROS). IEEE,  2018, pp. 5801--5806.

\bibitem[Shi \em{et~al.}(2015)Shi, Chen, Wang, Yeung, Wong, and
  Woo]{shi2015convolutional}
Shi, X.; Chen, Z.; Wang, H.; Yeung, D.Y.; Wong, W.K.; Woo, W.c.
\newblock Convolutional LSTM network: A machine learning approach for
  precipitation nowcasting.
\newblock {\em Advances in neural information processing systems} {\bf 2015},
  {\em 28}.

\bibitem[Aigner and K{\"o}rner(2018)]{aigner2018futuregan}
Aigner, S.; K{\"o}rner, M.
\newblock Futuregan: Anticipating the future frames of video sequences using
  spatio-temporal 3d convolutions in progressively growing gans.
\newblock {\em arXiv preprint arXiv:1810.01325} {\bf 2018}.

\bibitem[Lotter \em{et~al.}(2015)Lotter, Kreiman, and
  Cox]{lotter2015unsupervised}
Lotter, W.; Kreiman, G.; Cox, D.
\newblock Unsupervised learning of visual structure using predictive generative
  networks.
\newblock {\em arXiv preprint arXiv:1511.06380} {\bf 2015}.

\bibitem[Friston(2010)]{friston2010free}
Friston, K.
\newblock The free-energy principle: a unified brain theory?
\newblock {\em Nature reviews neuroscience} {\bf 2010}, {\em 11},~127--138.

\bibitem[Han and VanRullen(2017)]{han2017rhythms}
Han, B.; VanRullen, R.
\newblock The rhythms of predictive coding? Pre-stimulus phase modulates the
  influence of shape perception on luminance judgments.
\newblock {\em Scientific reports} {\bf 2017}, {\em 7},~1--10.

\bibitem[Kutas and Hillyard(1980)]{kutas1980reading}
Kutas, M.; Hillyard, S.A.
\newblock Reading senseless sentences: Brain potentials reflect semantic
  incongruity.
\newblock {\em Science} {\bf 1980}, {\em 207},~203--205.

\bibitem[Wang \em{et~al.}(2004)Wang, Bovik, Sheikh, and
  Simoncelli]{wang2004image}
Wang, Z.; Bovik, A.C.; Sheikh, H.R.; Simoncelli, E.P.
\newblock Image quality assessment: from error visibility to structural
  similarity.
\newblock {\em IEEE transactions on image processing} {\bf 2004}, {\em
  13},~600--612.

\bibitem[Hore and Ziou(2010)]{hore2010image}
Hore, A.; Ziou, D.
\newblock Image quality metrics: PSNR vs. SSIM.
\newblock In Proceedings of the 2010 20th international conference on pattern
  recognition. IEEE,  2010, pp. 2366--2369.

\bibitem[Zhang \em{et~al.}(2018)Zhang, Isola, Efros, Shechtman, and
  Wang]{zhang2018unreasonable}
Zhang, R.; Isola, P.; Efros, A.A.; Shechtman, E.; Wang, O.
\newblock The unreasonable effectiveness of deep features as a perceptual
  metric.
\newblock In Proceedings of the IEEE conference on computer vision and pattern
  recognition,  2018, pp. 586--595.

\bibitem[{\v{C}}ad{\'\i}k \em{et~al.}(2012){\v{C}}ad{\'\i}k, Herzog, Mantiuk,
  Myszkowski, and Seidel]{vcadik2012new}
{\v{C}}ad{\'\i}k, M.; Herzog, R.; Mantiuk, R.; Myszkowski, K.; Seidel, H.P.
\newblock New measurements reveal weaknesses of image quality metrics in
  evaluating graphics artifacts.
\newblock {\em ACM Transactions on Graphics (TOG)} {\bf 2012}, {\em 31},~1--10.

\bibitem[Oliu \em{et~al.}(2018)Oliu, Selva, and Escalera]{oliu2018folded}
Oliu, M.; Selva, J.; Escalera, S.
\newblock Folded recurrent neural networks for future video prediction.
\newblock In Proceedings of the European Conference on Computer Vision (ECCV),
  2018, pp. 716--731.

\bibitem[Wang \em{et~al.}(2017)Wang, Long, Wang, Gao, and Yu]{wang2017predrnn}
Wang, Y.; Long, M.; Wang, J.; Gao, Z.; Yu, P.S.
\newblock Predrnn: Recurrent neural networks for predictive learning using
  spatiotemporal lstms.
\newblock {\em Advances in neural information processing systems} {\bf 2017},
  {\em 30}.

\bibitem[Wang \em{et~al.}(2018{\natexlab{a}})Wang, Gao, Long, Wang, and
  Philip]{wang2018predrnn++}
Wang, Y.; Gao, Z.; Long, M.; Wang, J.; Philip, S.Y.
\newblock Predrnn++: Towards a resolution of the deep-in-time dilemma in
  spatiotemporal predictive learning.
\newblock In Proceedings of the International Conference on Machine Learning.
  PMLR,  2018, pp. 5123--5132.

\bibitem[Wang \em{et~al.}(2018{\natexlab{b}})Wang, Jiang, Yang, Li, Long, and
  Fei-Fei]{wang2018eidetic}
Wang, Y.; Jiang, L.; Yang, M.H.; Li, L.J.; Long, M.; Fei-Fei, L.
\newblock Eidetic 3d lstm: A model for video prediction and beyond.
\newblock In Proceedings of the International conference on learning
  representations,  2018.

\bibitem[Su \em{et~al.}(2020)Su, Byeon, Kossaifi, Huang, Kautz, and
  Anandkumar]{su2020convolutional}
Su, J.; Byeon, W.; Kossaifi, J.; Huang, F.; Kautz, J.; Anandkumar, A.
\newblock Convolutional tensor-train lstm for spatio-temporal learning.
\newblock {\em Advances in Neural Information Processing Systems} {\bf 2020},
  {\em 33},~13714--13726.

\bibitem[Lee \em{et~al.}(2021)Lee, Kim, Choi, Kim, and Ro]{lee2021video}
Lee, S.; Kim, H.G.; Choi, D.H.; Kim, H.I.; Ro, Y.M.
\newblock Video prediction recalling long-term motion context via memory
  alignment learning.
\newblock In Proceedings of the IEEE/CVF Conference on Computer Vision and
  Pattern Recognition,  2021, pp. 3054--3063.

\bibitem[Liu \em{et~al.}(2017)Liu, Yeh, Tang, Liu, and Agarwala]{liu2017video}
Liu, Z.; Yeh, R.A.; Tang, X.; Liu, Y.; Agarwala, A.
\newblock Video frame synthesis using deep voxel flow.
\newblock In Proceedings of the Proceedings of the IEEE International
  Conference on Computer Vision,  2017, pp. 4463--4471.

\bibitem[Wang \em{et~al.}(2018)Wang, Liu, Zhu, Liu, Tao, Kautz, and
  Catanzaro]{wang2018video}
Wang, T.C.; Liu, M.Y.; Zhu, J.Y.; Liu, G.; Tao, A.; Kautz, J.; Catanzaro, B.
\newblock Video-to-Video Synthesis.
\newblock {\em Conference on Neural Information Processing Systems (NeurIPS)}
  {\bf 2018}.

\bibitem[Wu \em{et~al.}(2020)Wu, Gao, Park, and Chen]{wu2020future}
Wu, Y.; Gao, R.; Park, J.; Chen, Q.
\newblock Future video synthesis with object motion prediction.
\newblock In Proceedings of the IEEE/CVF Conference on Computer Vision and
  Pattern Recognition,  2020, pp. 5539--5548.

\bibitem[Corlett \em{et~al.}(2009)Corlett, Frith, and
  Fletcher]{corlett2009drugs}
Corlett, P.R.; Frith, C.D.; Fletcher, P.C.
\newblock From drugs to deprivation: a Bayesian framework for understanding
  models of psychosis.
\newblock {\em Psychopharmacology} {\bf 2009}, {\em 206},~515--530.

\bibitem[Fletcher and Frith(2009)]{fletcher2009perceiving}
Fletcher, P.C.; Frith, C.D.
\newblock Perceiving is believing: a Bayesian approach to explaining the
  positive symptoms of schizophrenia.
\newblock {\em Nature Reviews Neuroscience} {\bf 2009}, {\em 10},~48--58.

\bibitem[Pearl(2009)]{pearl2009causality}
Pearl, J.
\newblock {\em Causality}; Cambridge university press,  2009.

\end{thebibliography}

%


\end{adjustwidth}
\end{document}